\pdfoutput=1

\documentclass[11pt]{article}

\usepackage{acl}

\usepackage{times}
\usepackage{latexsym}
\usepackage{graphicx} 
\usepackage{caption}
\usepackage{amsmath}
\usepackage{subcaption}

\usepackage[T1]{fontenc}

\usepackage[utf8]{inputenc}

\usepackage{microtype}

\usepackage{inconsolata}
\usepackage{multirow}
\usepackage{amsmath}

\title{HiGen: Hierarchy-Aware Sequence Generation for \\ Hierarchical Text Classification}

  \author{
 \hypersetup{linkcolor=black}
 Vidit Jain\thanks{\ \ Equal contribution}$^{*\clubsuit}$,
 Mukund Rungta\thanks{\ \ Work done by the author while at Georgia Institute of Technology}     $^{*\clubsuit\diamondsuit}$, 
  Yuchen Zhuang$^{\clubsuit}$, 
  Yue Yu$^{\clubsuit}$, 
  Zeyu Wang$^{\Phi}$,
  Mu Gao$^{\clubsuit}$, \\ 
  \hypersetup{linkcolor=black} \textbf{Jeffrey Skolnick}$^{\clubsuit}$, 
  \textbf{Chao Zhang}$^{\clubsuit}$ 
  \\
$^{\clubsuit}${Georgia Institute of Technology}, $^{\Phi}${Yale University} \\
$^{\diamondsuit}${Microsoft, Cambridge, USA}\\
 \textcolor{darkblue}{\texttt{\{\href{mailto:vjain312@gatech.edu}{vjain312},\href{mailto:mrungta8@gatech.edu}{mrungta8},\href{mailto:yczhuang@gatech.edu}{yczhuang},\href{mailto:yueyu@gatech.edu}{yueyu},\href{mailto:mu.gao@gatech.edu}{mu.gao},\href{mailto:skolnick@gatech.edu}{skolnick},\href{mailto:chaozhang@gatech.edu}{chaozhang}\}@gatech.edu}},
  \\
 \texttt{\href{mailto:zeyu.wang.zw389@yale.edu}{zeyu.wang.zw389@yale.edu}} \\
}

\begin{document}
\maketitle
\begin{abstract}

Hierarchical text classification (HTC) is a complex subtask under multi-label text classification, characterized by a hierarchical label taxonomy and data imbalance. The best-performing models aim to learn a static representation by combining document and hierarchical label information. However, the relevance of document sections can vary based on the hierarchy level, necessitating a dynamic document representation. To address this, we propose HiGen, a text-generation-based framework utilizing language models to encode dynamic text representations. We introduce a level-guided loss function to capture the relationship between text and label name semantics. Our approach incorporates a task-specific pretraining strategy, adapting the language model to in-domain knowledge and significantly enhancing performance for classes with limited examples. Furthermore, we present a new and valuable dataset called ENZYME, designed for HTC, which comprises articles from PubMed with the goal of predicting Enzyme Commission (EC) numbers. Through extensive experiments on the ENZYME dataset and the widely recognized WOS and NYT datasets, our methodology demonstrates superior performance, surpassing existing approaches while efficiently handling data and mitigating class imbalance. We release our code and dataset here: \url{https://github.com/viditjain99/HiGen}.

\end{abstract}

\section{Introduction}
Hierarchical text classification (HTC) is a task that involves categorizing text data into predefined categories organized in a hierarchical structure \cite{banerjee-etal-2019-hierarchical,wang-etal-2022-incorporating,zhou-etal-2020-hierarchy}. It holds great importance in various text mining applications, including scientific paper recommendation \cite{zhang2020multi,xu2023weakly}, semantic indexing \cite{li2019hiercon}, and online advertising \cite{agrawal2013multi}. HTC poses unique challenges when compared to traditional text classification, as it deals with imbalanced data distributions and complex dependencies between multiple levels of categories within the hierarchy. The hierarchical structure is represented as a directed acyclic graph (DAG), which must be encoded in the predictive model along with the text to generate the final hierarchal label. However, the data imbalance becomes more pronounced as we move down the levels, presenting a significant challenge for HTC.

Existing methods for hierarchical text classification (HTC) can be categorized into three groups: global \cite{zhou-etal-2020-hierarchy, chen-etal-2021-hierarchy}, local \cite{shimura-etal-2018-hft, banerjee-etal-2019-hierarchical, wehrmann2018hierarchical} and generative \cite{risch2020hierarchical, yu2022constrained, huang2022exploring}. Local approaches predict each hierarchical level using independent classifiers. Global models, on the other hand, use a single classifier and incorporate hierarchical information into the loss function. Finally, generative approaches use text generation frameworks to model the hierarchical structure. Some prior works flatten the label structure, leading to an exponential increase in the number of classes and the loss of hierarchical dependency. Additionally, some models fail to capture the correlation between text and label name semantics. As a result, these approaches struggle to perform well on long-tailed classes with limited training data.

Harnessing the power of large pretrained language models (PLMs) to capture text-label correlation, we employ a transformer-based sequence-to-sequence (seq2seq) framework \cite{lewis-etal-2020-bart}. These models are pretrained on extensive text data, enabling them to encode transferable linguistic features across tasks \cite{liu2019linguistic}. Our model transforms HTC into a text generation problem, generating labels conditioned on the input text and previously generated labels while utilizing a hierarchical level-guided semantics framework. While \citet{chen-etal-2021-hierarchy} introduces a matching loss for hierarchy-aware text-label relationships, it neglects the level-dependency between the transformed text and label representations. Our approach incorporates a margin loss, aligning text semantics with positive labels at each taxonomic level while pushing negative semantics apart. Furthermore, we propose a token constraint loss during training to discourage undesired token generation.


Inspired by the success of pretraining tasks \cite{gururangan2020don}, we propose a task-specific pretraining strategy for our generative model. Our approach presents a notably simplified design in contrast to previous works which use complex encoder architectures or training paradigms \cite{zhou-etal-2020-hierarchy, deng-etal-2021-htcinfomax, jiang2022exploiting}. We propose a pretraining step that takes advantage of weak supervision to jointly model text representations and hierarchical label information. The encoder encodes the input document and masked hierarchical label, while the decoder regenerates the original label. The pretraining step leverages weakly labeled data from the same domain, generated using an LLM. This domain-specific knowledge, discussed in \citet{mueller2022label, gururangan2020don}, proves advantageous for addressing the data imbalance challenge commonly encountered in HTC datasets.


The major contributions of this work are:
\begin{itemize}
\itemsep0em 
    \item We propose a generation-based HTC framework that effectively captures document-label dependencies across levels using a level-guided semantic loss.
    \item We devise an efficient pretraining strategy that leverages in-domain data to align the language model with the target task and domain.
    \item Our approach demonstrates remarkable performance on classes with limited examples, surpassing prior works even with minimal training instances.
    \item We present ENZYME, a dataset of 30,523 full-text PubMed articles with Enzyme Commission (EC) numbers. It features a four-level single-path hierarchy, making it larger than any existing datasets for biomedical HTC. 
\end{itemize}

\section{Related Work}
HTC is a multi-label classification problem where the classification task is performed in a hierarchical manner. Most prior work in HTC can be categorized into three broad categories: local, global, and generative. In the local approach, each hierarchical level has its own classifier tailored to the unique classes at that level. Conversely, the global approach employs a unified classifier that encompasses all classes across every hierarchy level. The generative approach, a recent advancement in hierarchical text classification, capitalizes on the generative capabilities of language models to predict text labels. More details on each of the categories are explained below. 

\textbf{Local HTC} Local approaches in HTC use local classifiers at each level or class \cite{banerjee-etal-2019-hierarchical, wehrmann2018hierarchical, shimura-etal-2018-hft, peng2018large}. \citet{banerjee-etal-2019-hierarchical} initialize binary classifiers at lower levels with parameters from the parent classifier. \citet{wehrmann2018hierarchical} combine local and global losses to encode information within and across hierarchical levels. \citet{shimura-etal-2018-hft} address data imbalance with parameter transfer techniques. \citet{peng2018large} employs a Graph-CNN-based model with recursive regularization for deep hierarchical representations.

\textbf{Global HTC} 
Global approaches to HTC utilize a single classifier \cite{10.1145/2487575.2487644, wu-etal-2019-learning, mao-etal-2019-hierarchical, 8933476} to predict labels at different hierarchy levels. Early works consider parent-child dependencies \cite{10.1145/2487575.2487644, wu-etal-2019-learning, mao-etal-2019-hierarchical, 8933476}, while recent approaches focus on global label structure \cite{wang2021cognitive}, disjoint features \cite{zhang2022hcn}, label imbalance \cite{deng-etal-2021-htcinfomax}, prior hierarchy knowledge \cite{zhou-etal-2020-hierarchy}, and semantic matching \cite{chen-etal-2021-hierarchy}. These works learn text and hierarchy semantics separately, fusing them later. However, \cite{wang-etal-2022-incorporating} proposes a global approach using contrastive learning to learn a shared representation. Similarly, \cite{jiang2022exploiting} also aims for common representations but incorporates both local and global hierarchies. Lastly, \cite{wang-etal-2022-hpt} uses prompt-tuning and multilabel MLM to learn shared semantics. 

\textbf{Generative HTC} Early works on text generation for HTC \cite{yang2018sgm, risch2020hierarchical} use RNN and Transformer-based seq2seq models, with dynamic document representations outperforming static encoder methods. Recent approaches \cite{yu2022constrained, huang2022exploring} propose T5-based models \cite{raffel2020exploring}. The former addresses label inconsistency with DFS-based linearization and constrained decoding, while the latter captures dependencies using BFS-based linearization and hierarchical path-based attention. These methods overlook the dependencies between the label name from different levels and the document text. We overcome this by capturing the document-label-name dependency across different levels using the proposed level-guided semantic loss.

In this work, we explore HTC under the umbrella of a sequence generation framework using a BART-based model \cite{lewis-etal-2020-bart}. We harness the benefits of denoising autoencoder pretraining to imbue the model with a strong understanding of the hierarchical structure.  We also employ a well-designed objective function during supervised training for improved learning.

\section{Problem Definition}
Given an input text $X_i = \{x_1, x_2, ..., x_n\}$, HTC aims to classify the text into a subset $Y_i$ of label set $Y$. The label set $Y$ is arranged as a Directed Acyclic Graph (DAG), denoted by $H = (L, E)$. $L$ represents the set of nodes and $E$ represents the edges indicating the nodes’ parent-child relations. All the labels in $Y$ constitute the nodes in the above graph i.e. $L$. We use BFS \cite{bundy1984breadth} to flatten the hierarchical labels into a multi-level sequential string of label nodes. We define a special set of symbols $S = \{ /, \text{$<$$root$$>$}\}$ to demarcate special relations of the label hierarchy in the flattened sequence. To classify $X_i$, the proposed sequence generation model generates $Y_i = \{y_1, y_2, ...., y_k\}$, where $y_j \in \{Y + S\}.$ BFS-based linearization helps each $Y_i$ to correspond to one or more paths in root to leaf paths in $H$.
\section{Methodology}

This section presents the proposed generative framework based on the BART model for HTC. We first describe the pretraining regime aimed at learning robust joint text-label representations (Section \ref{pretraining_strategy}). Following this, supervised training is performed to generate hierarchical label sequences for input documents (Section \ref{supervised_training}). We also explain the proposed objective function used (Section \ref{loss_function}) for this phase that aims to capture text and label semantics jointly.

\subsection{Pretraining Strategy}
\label{pretraining_strategy}

\begin{figure}[h!]
  \centering
  \includegraphics[width=\linewidth]{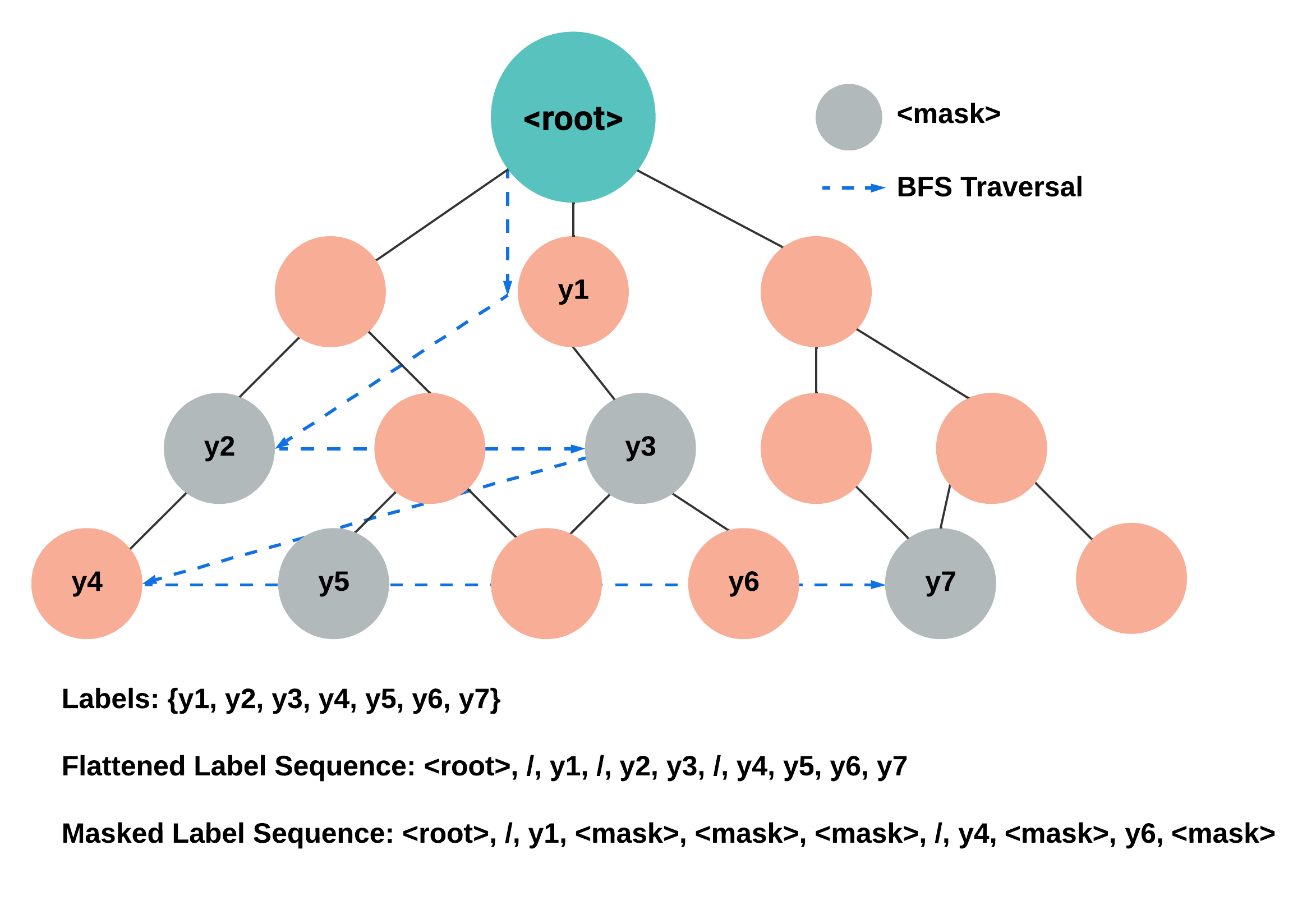}
  \caption{BFS-based label flattening and random token \& span masking employed during pretraining.}
  \label{fig:pretraining}
\end{figure}

Pretraining a domain-specific language model has shown significant performance improvement on downstream tasks \cite{gururangan2020don, mueller2022label}. We utilize the BART model which has achieved state-of-the-art results on text generation tasks,  as the backbone of our seq2seq approach. Its autoregressive nature and availability as a pre-trained model alleviate the reliance on extensive labeled data. The BART model comprises a transformer-based encoder and an autoregressive decoder.

To design our model, we adopt a BART-style denoising auto-encoder inspired by prior work \cite{aghajanyan2021htlm}. First, we transform a label set $Y_{i}$ into a multi-level sequential label $Y_{i}^{seq}$ using BFS, where $<$$root$$>$ represents the root node and $/$ indicates the change of level. Our approach involves randomly masking certain levels in the hierarchical label and encoding them with the input text. Figure \ref{fig:pretraining} provides an illustration of the label flattening and random masking techniques. The model is then trained to reconstruct the original hierarchical label. Formally, given a document $X_{i}$ and masked label $Y_{i}^{masked\_seq}$, we create an input sequence as follows:
\begin{equation}
\begin{aligned}
    input = [X_{i} \enspace \text{$<$$/s$$>$} \enspace Y_{i}^{masked\_seq}]
\end{aligned}
\end{equation}

where $<$$/s$$>$ is a special token used as a separator and $Y^{masked\_seq}_{i}$ is the masked label sequence resulting from the masking process as illustrated in Figure \ref{fig:pretraining}. We encode both the text and label using the same encoder, allowing the model to learn a joint embedding in the text-label space and capture correlations between them. During training, the model generates an output for the masked input sequence, aiming to fill in the masked positions of the label:
\begin{equation}
\begin{aligned}
    output = \hat{Y}^{seq}_{i}
\end{aligned}
\end{equation}

We use the cross-entropy loss, commonly used in masked language modeling, to train the model:
\begin{align*}
    loss = CE(Y_{i}^{seq}, \hat{Y}_{i}^{seq}) = -\sum_{i} Y_{i}^{seq} log(\hat{Y}_{i}^{seq}) 
\end{align*}

where $CE$ represents the cross-entropy loss, $\hat{Y}_{i}^{seq}$ represents the model's output probability distribution over the vocabulary.

The pretraining step aims to learn a joint representation for text and labels, capturing their inter-dependencies. The model is exposed to document-label pairs with partially masked hierarchical labels to develop a robust understanding of the hierarchy structure. Importantly, the model develops robust representations for all levels of the hierarchy, regardless of the number or levels of the masked nodes, as the extent of the masked label nodes is not known. The pretraining dataset is further described in Section \ref{pretraining_dataset}.

\subsection{Sequence to Sequence Modeling}
\label{supervised_training}
\begin{figure}[t!]
  \centering
  \includegraphics[width=\linewidth]{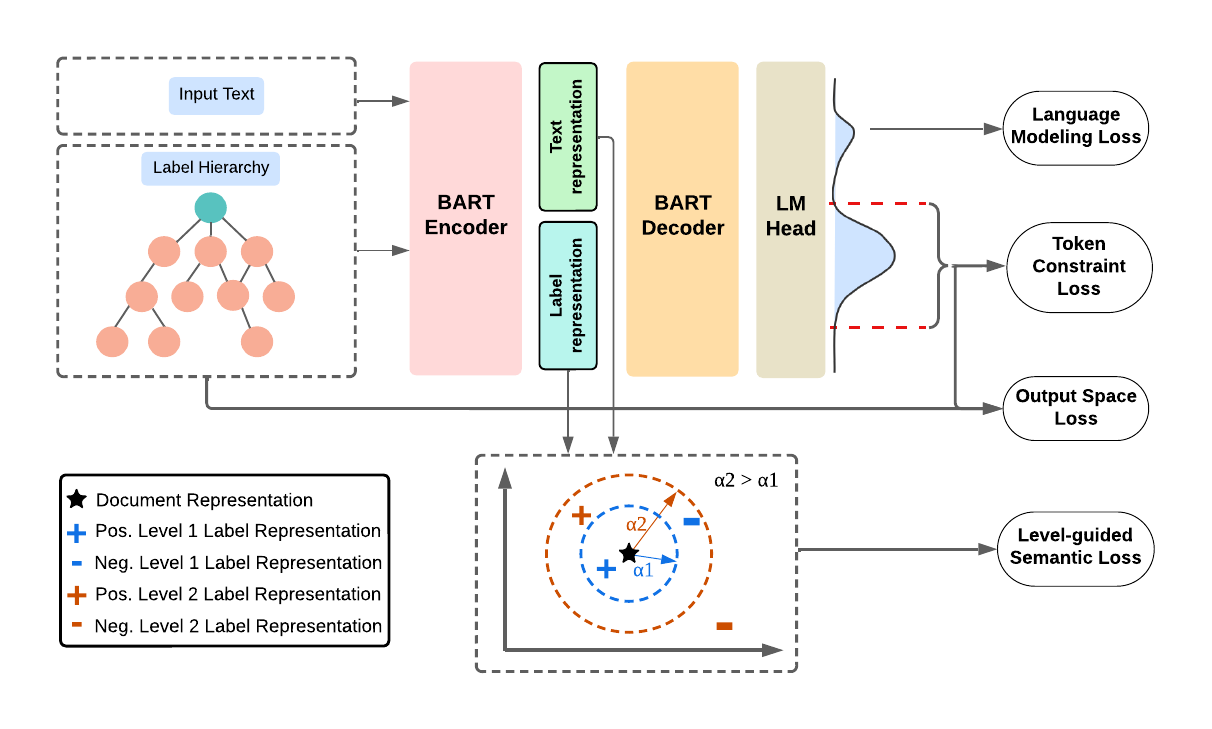}
  \caption{The proposed model's architecture consists of an encoder that takes the document and its corresponding label name as input. The decoder generates a hierarchical label output with 2 levels. To calculate the first three losses, the LM Head predicts a distribution over the vocabulary, and the hierarchy edges are considered. For the semantic loss, the text and label name representations from the encoder are projected onto a shared embedding space. Positive document and label semantics are pulled together, while negatives are pushed apart. The margins $\alpha_1$ and $\alpha_2$ control the attraction between levels 1 and 2, with $\alpha_1 > \alpha_2$.}
  \label{fig:main_model}
\end{figure}

The pretraining enhances the model's understanding of the label hierarchy and domain-specific knowledge. In the supervised training phase, we utilize the seq2seq framework for HTC, as shown in Figure \ref{fig:main_model}. We use the BFS label linearization to obtain multi-level label sequences. The input text document $X_i = \{x_1, x_2, x_3, \cdots, x_m \}$ is encoded to obtain the encoder output hidden representation $h^e$.
\begin{equation}
\begin{aligned}
    h^{e} = \text{Encoder}(X_i)
\end{aligned}
\end{equation}

The hidden representations from the encoder are used to initialize the decoder. The decoder then generates the hierarchical label $\hat{Y}_{i}$ step-by-step autoregressively. The autoregressive process followed by the decoder can be represented as:
\begin{equation}
\begin{aligned}
    p(\hat{Y}_{i} \mid X_{i})=\prod_{k=1}^n p\left(\hat{Y}^{k}_{i} \mid X_{i}, \hat{Y}_{i}^{<k}\right)
\end{aligned}
\end{equation}

where $\hat{Y}^{k}_{i}$ denotes the prediction from the decoder for level $k$ of the hierarchy, and $n$ represents the depth of the hierarchy. At every time step $k$, the output $\hat{Y}^{k-1}_{i}$ and the hidden state $h^{d}_{k-1}$ from the previous time step $k-1$ is given as input to the decoder to generate the next hidden state $h^{d}_{k}$ and a prediction $\hat{Y}^{k}_{i}$ for the current time step.
\begin{equation}
\begin{aligned}
    h^{d}_{k}, \hat{Y}^{k}_{i} = \text{Decoder}(h^e,h^{d}_{k-1}, \hat{Y}^{k-1}_{i} )
\end{aligned}
\end{equation}

The decoder leverages information from the encoded document and the label from the previous level to predict the current level in the hierarchy. This allows our model to mimic the HTC process and learn important aspects of the hierarchy structure, including label relations, valid root-to-leaf node paths, and the overall label structure \cite{risch2020hierarchical}.

\subsection{Training Objective}
\label{loss_function}
This section presents the proposed objective function we use to train our generative framework for the proposed HTC task. First, we provide a description of each loss function being employed and then define the final training objective used to train our model. 


\textbf{Language Modeling Loss.} The predictions from each time step of the decoder $\hat{Y}^{k}_{i}$ are concatenated together to form the final prediction of the model $\hat{Y}_{i}$. The ground truth, ${Y}_{i}$ comprises the original flattened label. The language modeling loss for HiGen can be expressed as:
\begin{equation}
\begin{aligned}
    \mathcal{L}_{LM} = \text{crossentropy}({Y}_{i}, \hat{Y}_{i})
\end{aligned}
\end{equation}

\textbf{Output Space Loss.} The label hierarchy is generally represented as a DAG, where each edge signifies a parent-to-child relationship. In HTC, imposing this unidirectionality during training helps the model comprehend these hierarchical relations. To this end, we use the formulation proposed by \citet{zhang2021match} inspired by the distributional inclusion hypothesis (DIH) \cite{geffet2005distributional}.
\begin{equation}
\begin{aligned}
    \mathcal{L}_{O} = \sum_{N}\sum_{l}\sum_{n} max(0, \pi_{c} - \pi_{p})
\end{aligned}
\end{equation}

where $N$ is the batch size, $l$ is the number of predicted node labels and $n$ is the number of hierarchy edges. $\pi_{p}$ and $\pi_{c}$ represent the predicted probabilities for a parent label node $p$ and its child label node $c$ in the output distribution over the vocabulary.

Under the DIH framework, the loss term can be precisely articulated as follows: if a document $d$ belongs to a child class $c$ with probability $\pi_{c}$, then it must belong to the parent class $p$ with a probability no less than $\pi_{c}$. For instance, if a document has a 75\% chance of being labeled with \textit{"Football"}, the likelihood of it being assigned to the parent category \textit{"Sports"} should be 75\% or higher. Note that the loss term exhibits asymmetry, being non-zero when $\pi_{c} > \pi_{p}$ but zero when $\pi_{p} > \pi_{c}$. Referring to Figure \ref{fig:main_model}, loss calculation involves aggregating predicted probabilities for valid tokens (indicated by red dotted lines) from the LM Head and establishing parent-child pairs using the hierarchy edges. Through the imposition of a penalty when $\pi_{c} > \pi_{p}$, the objective function guides the model to learn the correct sequence of tokens, consequently learning the unidirectionality and orientation of the hierarchy edges. 

\textbf{Token Constraint Loss.} Using a model trained for open-ended text generation in HTC can lead to irrelevant node labels due to the generation of stray tokens (non-existent in a given hierarchy). To address this and avoid restricting BART's generation capabilities due to a fixed vocabulary, instead, we introduce a loss function that penalizes predictions outside of a designated vocabulary represented by the red dotted lines in Figure \ref{fig:main_model}. This vocabulary is constructed using the label hierarchy, ensuring alignment and enhancing overall results. Putting it formally,
\begin{equation}
\begin{aligned}
    & V^{H'} = \{l \in V | l \notin V^{H}\} \\
    & \mathcal{L}_{T} = \sum_{l \in V^{H'}} \pi_{l}^{i}
\end{aligned}
\end{equation}

where $l$ is a token, $\pi^{i}$ represents a vector of predicted token probabilities for a training example by the LM Head, $V$ is the entire vocabulary, $V^{H}$ is the desired vocabulary for $H$, and $V^{H'}$ contains the remaining tokens. The loss function guides the model to learn the desired vocabulary's contents, discouraging the generation of irrelevant labels and enhancing classification performance.

\textbf{Level-guided Semantic Loss.} Based on the previous loss functions, the model learns correlations between text semantics and the hierarchical structure. Label names serve as descriptions encoding paths in the hierarchy, providing valuable semantic information for HTC. For any classification task, the text and label semantics for positive pairs should be closer than unrelated pairs in an embedding space. Our training objective imposes this constraint, guided by label nodes at each hierarchical level.

In a traditional classification setting, document similarity can be determined by matching their labels. However, in HTC, some documents might be related to each other for some initial levels of the hierarchy but then diverge as we move down and vice versa. Capturing this nuance is crucial for HTC and is the prime distinction from traditional classification. We aim to encode this information using our level-guided semantic loss function. 

We utilize two independent fully connected networks and project the text and label semantics onto a common embedding space.
\begin{equation}
\begin{aligned}
    E_{t} = FC_{t}(h_{t}^{e}) \\
    E_{l} = FC_{l}(h_{l}^{e})
\end{aligned}
\end{equation}

where $FC_{t}$ \& $FC_{l}$ are two fully-connected networks, $h_{t}^{e}$ \& $h_{l}^{e}$ are the encoder hidden representation for the document and label name. $E_{t}$, $E_{l} \in \mathcal{R}^{N \times d}$ represent the text and label name representations in the joint space and $N$ is the batch size.

To combine the hierarchical information with text and label semantics, we construct document-label name pairs for every level of the hierarchy. So for a batch of data and a particular level $k$ from the hierarchy, the pairing process involves associating each document embedding with every label name embedding in the batch. Pairs receive a positive label (1) when the document and label name embeddings correspond to the same level label; otherwise, a negative label (0) is assigned. Consequently, a positive document-label name embedding pair is denoted as $\{E_{t_{i}}^{+}, E_{l_{j}}^{+}\}$, while a negative pair is denoted as $\{E_{t_{i}}^{-}, E_{l_{j}}^{-}\}$. Intra-pair distances are calculated amongst the positive and negative document-label pairs using L2-normalized Euclidean distance. We use the mean of intra-document-label pair distances for both the positive and negative variants for further calculations. Mathematically, 
\begin{equation}
\begin{aligned}
    & \mathcal\gamma_{k}^{+} = \frac{1}{N^{+}} \sum_{i}\sum_{j} ||E^{+}_{t_{i}} - E^{+}_{l_{j}}|| \\
    & \mathcal\gamma_{k}^{-} = \frac{1}{N^{-}} \sum_{i}\sum_{j} ||E^{-}_{t_{i}} - E^{-}_{l_{j}}||
\end{aligned}
\end{equation}

To bring the positive document and label name semantics closer, a margin loss is applied with a level-specific margin parameter $\alpha_{k}$. So for the level $k$, the loss is defined as,
\begin{equation}
\begin{aligned}
    \mathcal{L}_{S}^{k} = \max(0, \gamma_{k}^{+} - \gamma_{k}^{-} + \alpha_{k})
\end{aligned}
\end{equation}

At every hierarchy level, we replicate the pairing and loss calculation using the outlined strategy. Notably, with level transitions, documents initially labeled as negatives may now share labels and vice versa, creating a fresh set of labeled pairs. The level-specific margin $\alpha_{k}$ progressively increases as we descend the hierarchy, to incorporate the increasing granularity of labels. This ensures the model aligns text and label semantics effectively. For a label hierarchy with $m$ levels, the final loss becomes,
\begin{equation}
\begin{aligned}
    \mathcal{L}_{S} = \sum_{k=1}^{m} \mathcal{L}_{S}^{k}
\end{aligned}
\end{equation}

In Figure \ref{fig:main_model}, the lower part illustrates an example of a two-level hierarchy, with different colors representing the hierarchy levels. For level 1, positive label semantics are attracted to document semantics, while negative label semantics are pushed apart by at least $\alpha_1$ (blue-dotted circle). The same process is applied to the next level, with increased separation $\alpha_2$ to accommodate the higher semantic granularity (orange-dotted circle).

Based on the language modeling loss and the three proposed loss functions, we use the following objective function to learn the parameters of our model:
\begin{align*}
    \min\text{ }\mathbf{\mathcal{L}_{HiGen}} = \mathcal{L}_{LM} + \lambda_{1}\mathcal{L}_{O} + \lambda_{2}\mathcal{L}_{T} + \lambda_{3}\mathcal{L}_{S}
\end{align*}

where, $\lambda_1$, $\lambda_2$ and $\lambda_3$ are hyperparamters.

\section{Experiments}
\subsection{Datasets}

\begin{table}[h!]
    \centering
    \resizebox{\columnwidth}{!}{
    \begin{tabular}{lcccccc}
        \hline
        Dataset & $|L|$ & Depth & $Avg(|L_i|)$ & Train & Val & Test \\
        \hline
        ENZYME & 4566 & 4 & 4 & 17422 & 8741 & 4360  \\
        WOS & 141 & 2 & 2 & 30070 & 7518 & 9397  \\
        NYT & 166 & 8 & 7.6 & 23345 & 5834 & 7292 \\
        \hline
    \end{tabular}
    }
    \caption{\label{tab:data_statistics} Statistics of datasets used. $|L|$: total number of target classes, Depth: levels of the hierarchy, $Avg(|L_i|)$: Average number of classes per example.
    }
    \vspace{-1em}
\end{table}

We introduce the \textbf{ENZYME} dataset, containing biomedical scientific literature and corresponding Enzyme Commission (EC) numbers. We provide an overview of this dataset, including the training, validation, and test splits, in Section \ref{enzyme_dataset}. We also conduct experiments on benchmark datasets: Web-of-Science (WOS) \cite{8260658} and NYT \cite{sandhaus2008new}, following the preprocessing and data splits proposed by \citet{zhou-etal-2020-hierarchy}. WOS and ENZYME focus on single-path HTC, while NYT incorporates multi-path taxonomic labels. Detailed statistical information can be found in Table \ref{tab:data_statistics}. Experimental results are evaluated using Macro-F1 and Micro-F1 metrics, commonly used in prior literature.

\subsubsection{Pretraining Datasets} 
\label{pretraining_dataset}
For the ENZYME dataset, we use the articles extracted from PubMed\footnote{\url{https://pubmed.ncbi.nlm.nih.gov/advanced/}}. We randomly sample 200,000 articles along with their Enzyme Commission (EC) numbers that follow a hierarchical structure. These articles are loosely labeled as they do not reflect human-annotated EC numbers. WOS hosts a comprehensive collection of scientific articles spanning various domains of science. To produce meaningful and diverse abstracts, we employ the powerful ChatGPT model and generate $\sim$3000 abstracts for pretraining. More details are mentioned in Appendix \ref{sec:chatgpt_pretraining}. As for NYT, we have access to a vast repository of articles that were not assigned to specific training, testing, or validation sets by \citet{zhou-etal-2020-hierarchy}. This invaluable resource enables us to utilize these uncategorized articles for our proposed pretraining task. During the process of consolidating the pretraining dataset, we made sure to prevent any overlap of the pretraining data with the training, validation and testing sets.

\begin{table*}[!ht]
    \centering
    \resizebox{0.9\linewidth}{!}{%
    \begin{tabular}{ccccccc}
    \hline
    \multirow{2}{*}{Models} & \multicolumn{2}{c}{ENZYME} & \multicolumn{2}{c}{WOS} & \multicolumn{2}{c}{NYT} \\ 
    \cline{2-7}
    & Micro-F1 & Macro-F1 & Micro-F1 & Macro-F1 & Micro-F1 & Macro-F1  \\ \hline
        BERT & 82.16 & 16.29 & 85.63$^*$ & 79.07$^*$ & 78.24$^*$ & 65.62$^*$ \\ 
        HiAGM \cite{zhou-etal-2020-hierarchy} & 80.52 & 49.21 & 85.82 & 80.28 & 74.97 & 60.83 \\ 
        HiMatch \cite{chen-etal-2021-hierarchy} & 72.37 & 38.19 & 86.20 & 80.53 & 74.62 & 59.28 \\
        HiMatch + BERT & 75.24 & 40.67 & 86.70$^*$ & 81.06$^*$ & 76.79$^*$ & 63.89$^*$ \\ 
        HTCInfoMax \cite{deng-etal-2021-htcinfomax} & 70.56 & 37.24 & 85.58 & 80.05 & - & - \\ 
        HTCInfoMax + BERT & 73.45 & 39.83 & 86.30$^*$ & 79.97$^*$ & 78.75$^*$ & 67.31$^*$ \\ 
        HGCLR \cite{wang-etal-2022-incorporating} & 90.81 & 71.03 & 87.11 & 81.20 & 78.86 & 67.96 \\ 
        HPT \cite{wang-etal-2022-hpt} & 91.04 & 78.27 & 87.16 & 81.93 & 80.42 & 70.42 \\
        HBGL \cite{jiang2022exploiting} & 91.10 & 83.05 & 87.36 & 82.00 & 80.47 & 70.19 \\ 
        \hline
        Vanilla BART & 88.11 & 67.76 & 86.26 & 79.34 & 80.08 & 69.3 \\
        SGM-T5$^\dag$ (from \citealp{yu2022constrained}) & - & - & 85.83 & 80.79 & - & - \\ 
        Seq2Tree-T5$^\dag$ \cite{yu2022constrained} & - & - & 87.20 & \textbf{82.50} & - & - \\ 
        PAAM-HiA-T5$^\dag$ \cite{huang2022exploring} & - & - & \textbf{90.36} & 81.64 & 77.52 & 65.97 \\ 
        HiGen (ours) & \textbf{92.61} & \textbf{84.15} & 87.39 & 81.45 & \textbf{80.89} & \textbf{72.41} \\ \hline
    \end{tabular}}
    \caption{\label{tab:comparison}Experimental results of our proposed approach on all datasets. $\dag$: Implementation not available, $*$: results from \citet{wang-etal-2022-incorporating}.
    }
    \vspace{-0.5em}
    
\end{table*}

\subsubsection{ENZYME Dataset}
\label{enzyme_dataset}

We introduce a new dataset called ENZYME, which contains curated full-text biomedical articles from PubMed along with Enzyme Commission (EC) numbers (see Section \ref{ec_system}) and enzyme names. It consists of 30,523 articles in both PDF and parsed formats, making it unique in providing full-text biomedical documents with corresponding enzyme identification numbers. The EC numbers follow a hierarchical structure\footnote{\url{https://www.enzyme-database.org/contents.php}}, with detailed statistics per level shown in Table \ref{tab:classes_per_level}. The dataset has a hierarchical taxonomy depth of 4, resulting in a complex structure with a large number of fine-grained classes, making this an unique and challenging dataset. This dataset is highly imbalanced with $\sim$50\% of the 4,566 classes having less than 2 examples (Figure \ref{fig:data_statistics}). For our experiments, we focus on classes with 5 or more examples. Further details on the enzyme classification system and dataset construction are available in Appendix \ref{sec:enzyme_appendix}.

\subsubsection{Baselines}
To compare the proposed method, we select a few recent baselines. HBGL \cite{jiang2022exploiting}, SGM-T5 \cite{yang2018sgm}, PAAM-HiA-T5 \cite{huang2022exploring}, Seq2Tree-T5 \cite{yu2022constrained}, HGCLR \cite{wang-etal-2022-incorporating}, HPT \cite{wang-etal-2022-hpt}, HiMatch \cite{chen-etal-2021-hierarchy} and HiAGM \cite{zhou-etal-2020-hierarchy}, and HTCInfoMax \cite{deng-etal-2021-htcinfomax}. SGM-T5, PAAM-HiA-T5 and Seq2Tree-T5 have used a similar sequence generation approach for HTC. 
SGM uses a T5 encoder-decoder framework, while Seq2Tree-T5 proposes a tree-like framework with label linearization. HBGL leverages BERT's large-scale parameters and language knowledge to model global and local hierarchies. HiAGM, HTCInfoMax and HiMatch incorporate fusion strategies to integrate text and hierarchy representations. HiAGM introduces hierarchy-aware multi-label attention, HTCInfoMax employs information maximization for modeling text-hierarchy interaction, and HiMatch matches text and label representations in a joint embedding space for classification. We also compare our approach with Vanilla-BART and a BERT-based HTC model.


\subsubsection{Implementation Details}


For HiGen, we use HuggingFace checkpoints to warm start the BART model, selecting between a PubMed-finetuned BART model (mse30/bart-base-finetuned-pubmed) for ENZYME and the base-BART model (facebook/bart-base) for WOS and NYT datasets. After an initial pretraining phase using the data in Section \ref{pretraining_dataset}, we save the checkpoints and subsequently fine-tune them for all datasets. For WOS, we set semantic margins $\alpha_k$ to [0.05, 0.1], while for ENZYME, they are [0.02, 0.1, 0.15, 0.3]. Due to label sequence complexity, semantic loss is not applied to NYT at present. We use loss balancing factors $\lambda_1$ and $\lambda_2$ of [1e-3, 1e-6] for ENZYME, and [1e-3, 1e-5] for WOS and NYT, with $\lambda_3$ set to 1 for ENZYME and WOS. These values result from extensive hyperparameter tuning. Both pretraining and fine-tuning use a batch size of 12, the Adam optimizer, and a learning rate of 5e-5 with a linear schedule.

We employ consistent evaluation across all datasets by using the baseline implementations provided by their respective authors. For BERT, we train a multilabel classification model using the representation of the special $[CLS]$ token. Like HiGen, BERT is warm started with a PubMed-finetuned checkpoint (microsoft/BiomedNLP-PubMedBERT-base-uncased-abstract-fulltext) for ENZYME, and base-BERT (bert-base-uncased) for WOS and NYT. All models are implemented in PyTorch.


\subsection{Experimental Results}



Table \ref{tab:comparison} presents the Micro and Macro F1 scores for the baselines and HiGen on the three datasets. The baselines are grouped into two categories: classification-based and generation-based. HiGen, our generation-based approach, achieves superior performance over both categories.
Among the classification-based methods, HBGL stands out as the state-of-the-art approach, leveraging global and local hierarchies for text and label representation. To ensure a fair comparison on ENZYME, we initialize HiMatch + BERT, HGCLR, and HBGL with the PubMed BERT checkpoint. HiGen, utilizing dynamic text and label semantics, improves the Micro-F1 score by 1.51\%, 0.03\%, and 0.42\% for ENZYME, WOS, and NYT respectively. Notably, HiGen exhibits exceptional performance on the ENZYME dataset for both Micro and Macro-F1 scores. A significant improvement over the Vanilla-BART establishes the power of proposed pertaining and loss functions. Larger improvement can be seen for dataset involving deeper hierarchy (ENZYME and NYT) compared to shallower hierarchy (WOS).

HiGen uses the BART model (140 million parameters) instead of the T5 model (220 million parameters), making it 37\% more parameter-efficient. HiGen outperforms Seq2Tree-T5 for WOS. For NYT, HiGen surpasses PAAM-HiA-T5 by 3.47\% and 6.44\% on Micro and Macro-F1 scores, respectively. Performance figures on ENZYME are not available due to code unavailability. 

In Appendix \ref{computational_cost}, we juxtapose the training durations of HiGen and HBGL, revealing a 10-fold reduction in training time for HiGen, attributable to its simpler architecture, as evidenced in Table \ref{tab:training_times}.

\subsection{Performance on Long-Tailed Distribution}
\label{long_tailed}
\begin{figure}[t!]
     \centering
     \begin{subfigure}[b]{0.49\linewidth}
         \centering
         \includegraphics[width=\linewidth]{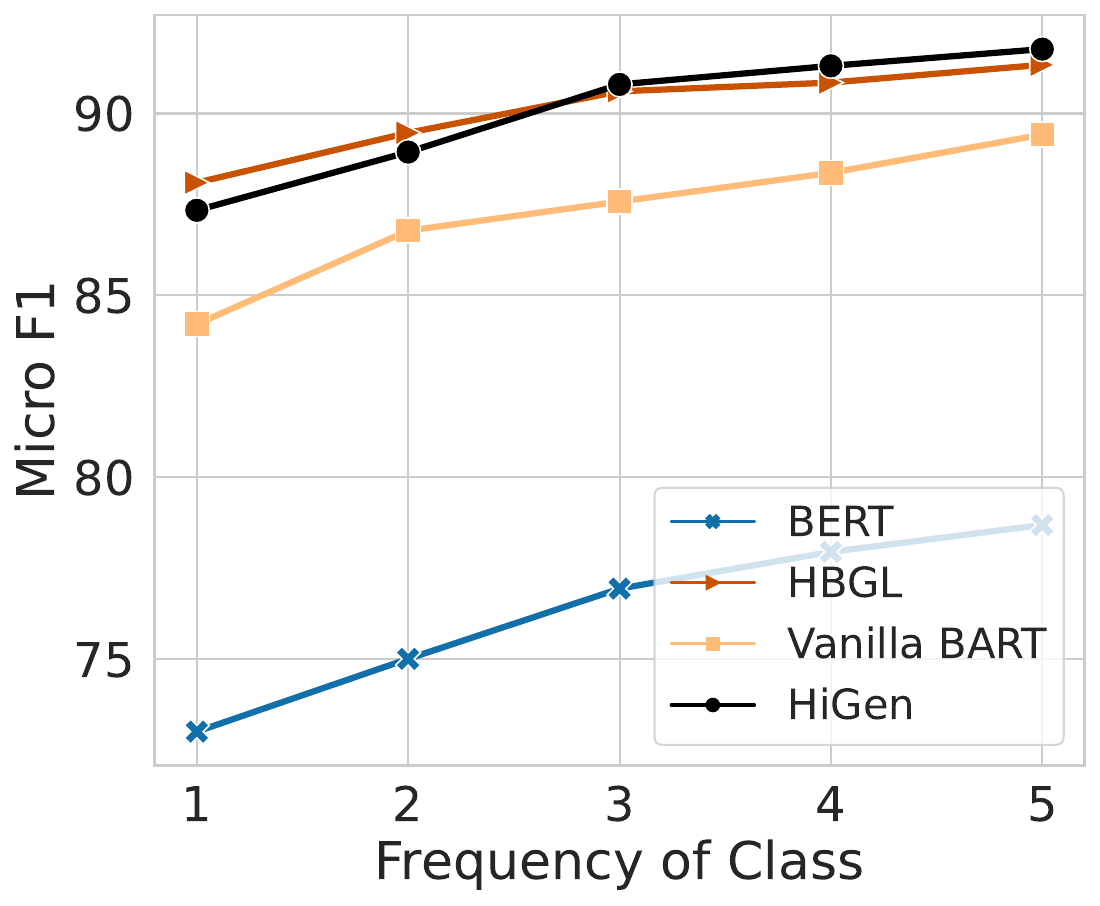}
         \caption{Micro-F1 v/s Frequency of Classes}
         \label{fig:longtailed_micro}
     \end{subfigure}
     \hfill
     \begin{subfigure}[b]{0.49\linewidth}
         \centering
         \includegraphics[width=\linewidth]{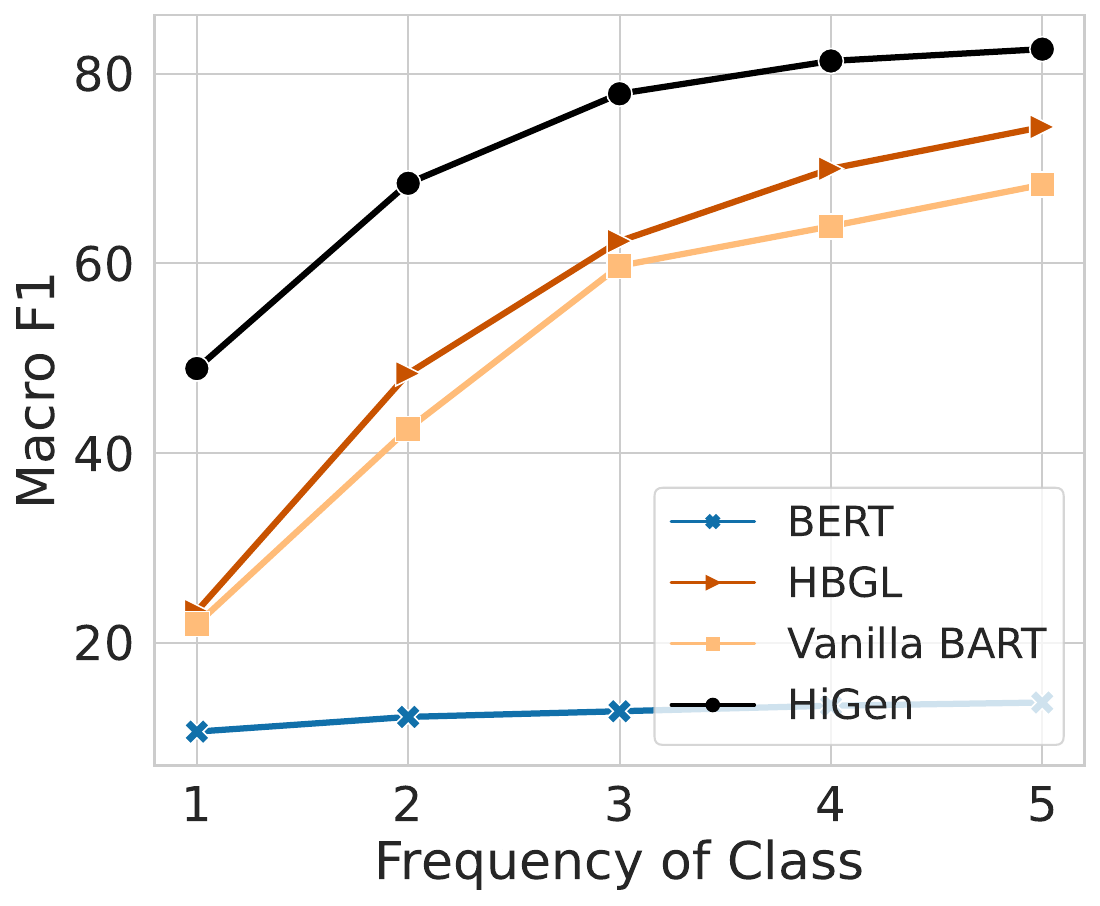}
         \caption{Macro-F1 v/s Frequency of Classes}
         \label{fig:longtailed_macro}
     \end{subfigure}
     \caption{Performance on ENZYME dataset for the long-tailed classes}
     \label{fig:longtailed}
     \vspace{-1.5em}
\end{figure}

In our experiments, data imbalance is evident, with \emph{long-tailed classes} having limited training and testing examples. These classes are crucial for evaluating the model's performance, testing its ability to learn from sparse data, and assessing generalization capabilities. We select classes based on testing data frequency, grouping them into bins from 1 to 5.

Figure \ref{fig:longtailed} presents results for the ENZYME dataset, displaying Micro-F1 (Figure \ref{fig:longtailed_micro}) and Macro-F1 (Figure \ref{fig:longtailed_macro}) scores against class frequency. We compare HiGen to baseline models BERT, HBGL, and Vanilla BART, with a focus on HBGL as the best-performing baseline.  HiGen outperforms BERT and Vanilla BART significantly in both Micro and Macro-F1. While its Micro-F1 is similar to HBGL, HiGen excels in Macro-F1, particularly for "long-tailed classes" on the ENZYME dataset.

\subsection{Data Efficiency}
\label{data_efficiency}
\begin{figure}[t!]
     \centering
     \begin{subfigure}[b]{0.49\linewidth}
         \centering
         \includegraphics[width=\linewidth]{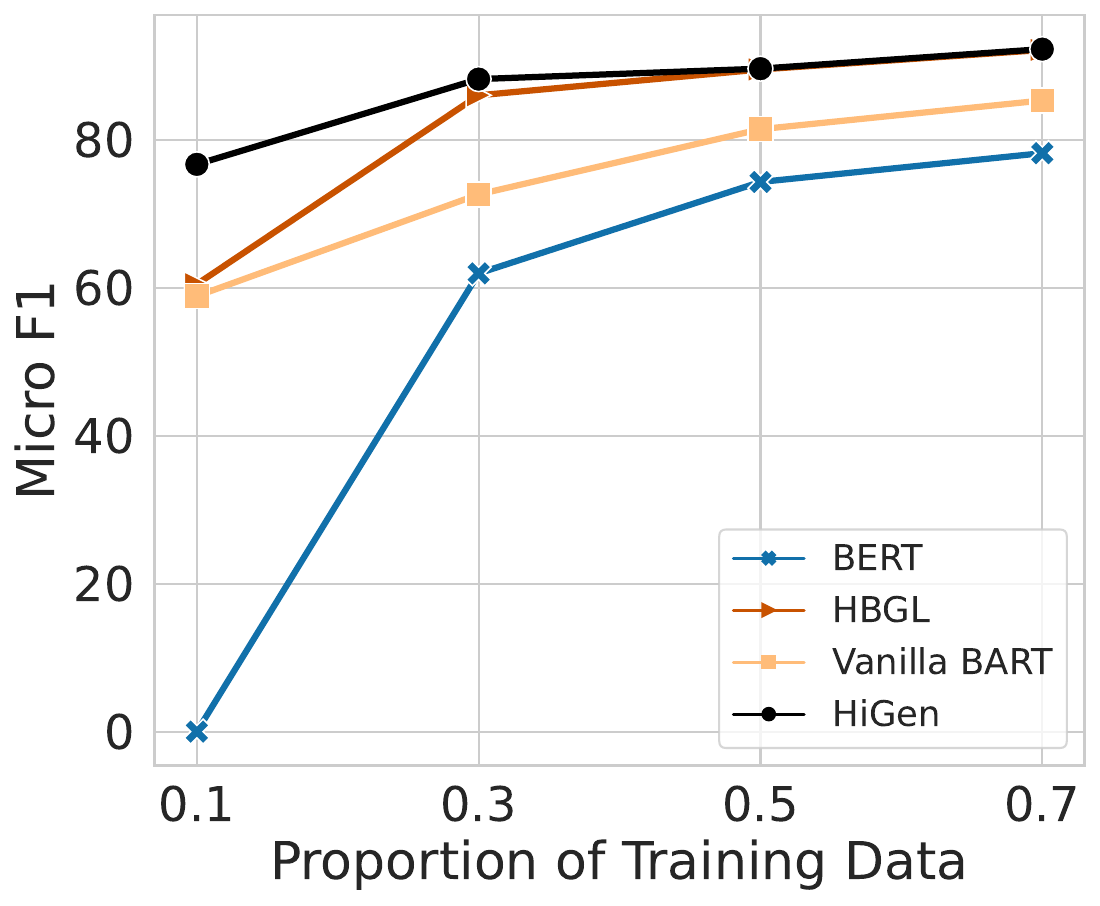}
         \caption{Micro-F1 v/s Data proportion}
         \label{fig:limited_micro}
     \end{subfigure}
     \hfill
     \begin{subfigure}[b]{0.49\linewidth}
         \centering
         \includegraphics[width=\linewidth]{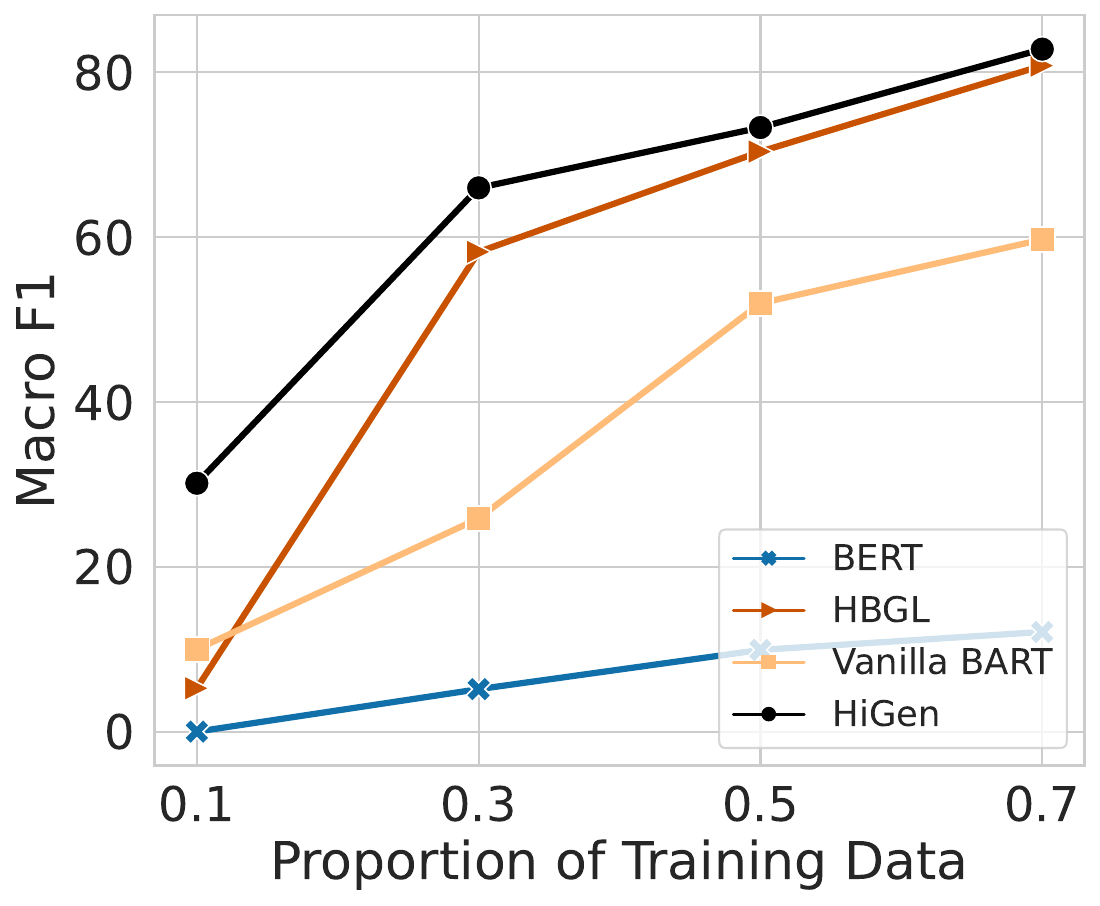}
         \caption{Macro-F1 v/s Data proportion}
         \label{fig:limited_macro}
     \end{subfigure}
     \caption{Performance on ENZYME dataset on varying the training data proportion}
     \label{fig:limited}
     \vspace{-0.5em}
\end{figure}

We assessed model robustness by training on different data proportions and evaluating on the original test set. We included baseline models like BERT, Vanilla BART, and HBGL for fair comparisons. Figure \ref{fig:limited} displays results for the ENZYME dataset, known for its complex hierarchy and limited training examples. The curves represent model performance as training data increases.

Our model outperformed all baseline models with just 10\% of the data, surpassing the second-best model, HBGL, thanks to pretraining knowledge. This synergy of pretraining and fine-tuning improved label hierarchy knowledge, enhancing performance. For more details, see Section \ref{ablation_study}.

\subsection{Ablation Study}

\label{ablation_study}


In this analysis, we evaluate the impact of individual components in our approach on the WOS dataset (Table \ref{tab:ablation}). The importance of the pretraining stage is evident, with its removal leading to a significant drop in performance across both metrics. Even with just 3000 abstracts generated using LLMs like ChatGPT, our pretraining approach is highly effective, emphasizing their value for data-limited datasets. By refining prompts and adding more pretraining data, we expect further performance improvements. Both the semantic loss ($\mathcal{L}_S$) and token constraint loss ($\mathcal{L}_T$) significantly improve the Macro-F1 score by enhancing representation learning for minority classes. While the output space loss ($\mathcal{L}_O$) also contributes to Macro-F1 improvement, it has a comparatively smaller impact. In summary, all proposed components are crucial; removing any of them leads to a notable drop in performance, as evident when comparing HiGen to a Vanilla BART model. For further results and analyses, please refer to Appendix \ref{sec:ablation_study_2}.

\begin{table}[t!]
    \centering
    \resizebox{0.7\linewidth}{!}{%
    \begin{tabular}{lcc}
    \hline
    Model & Micro-F1 & Macro-F1\\
    \hline
    Vanilla BART & 86.26 & 79.34 \\
    HiGen & \textbf{87.39} & \textbf{81.45} \\
    \hspace{2mm}w/o pretraining & 86.69 & 80.54 \\
    \hspace{2mm}w/o $\mathcal{L}_O$ & 87.35 & 81.11 \\
    \hspace{2mm}w/o $\mathcal{L}_T$ & 87.19 & 80.83 \\
    \hspace{2mm}w/o $\mathcal{L}_S$ & 87.33 & 80.91 \\
    \hline
\end{tabular}}
\caption{\label{tab:ablation} Ablation study on HiGen for the WOS dataset.}
\vspace{-1.5em}
\end{table}


\section{Conclusion}
In this paper, we propose a sequence generation framework for HTC that captures the hierarchical nature of labels. The pretraining strategy enhances the model's performance by adapting it to the specific domain and task. Moreover, we demonstrate the effectiveness of utilizing synthetic data generated through powerful language models like ChatGPT. The objective function during supervised training provides the model with additional contextual information about the hierarchical structure. Our proposed approach outperforms baseline models on three datasets, as measured by standard evaluation metrics. Leveraging the knowledge embedded in pretrained language models, our model performs exceptionally well on classes with limited examples and is data-efficient as well.
\section*{Limitations}
While HiGen demonstrates performance improvement across three datasets, it is crucial to acknowledge certain limitations and identify potential areas for future improvement. 

Firstly, ensuring the appropriate nature of the pretraining data is paramount. It is essential that the data aligns with the same domain as the original dataset, and more importantly, that the labels in the pretraining data adhere to the same hierarchical structure as the original data. If such data is not readily available, alternative sources like large language models for obtaining weakly supervised pretraining data may need to be utilized.

Secondly, in the level-guided semantic loss, the current approach employs in-batch sampling. However, the selection of the positive and negative samples in a batch might not be optimal. Future works can explore enhancements by incorporating harder negative samples, which could potentially improve the model's ability to learn more discriminative representations.

Lastly, the present approach involves several hyperparameters which can introduce additional complexity and computational overhead during the fine-tuning process. Future research efforts could focus on streamlining and simplifying these aspects to ensure a more efficient and user-friendly implementation.

\section*{Acknowledgement}
This work was supported in part by NSF IIS-2008334, IIS-2106961, and CAREER IIS-2144338.

\bibliography{custom}




\appendix

\section{ENZYME Dataset}
\label{sec:enzyme_appendix}

\begin{figure}[!ht]
  \centering
  \includegraphics[width=0.7\linewidth]{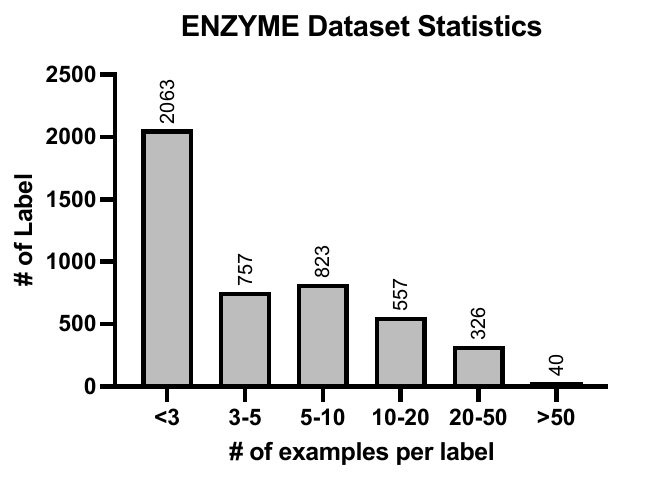}
  \caption{ENZYME dataset statistics}
   \label{fig:data_statistics}
\end{figure}

\subsection{Enzyme Classification (EC) System}
\label{ec_system}
The Enzyme Commission (EC) number is a system for classifying enzymes, based on the chemical reactions they catalyze. The number is made up of four digits separated by periods and shows the enzyme's class, subclass, sub-subclass, and serial number respectively. This number is used to identify enzymes uniquely in the Enzyme Nomenclature database managed by the International Union of Biochemistry and Molecular Biology (IUBMB). The first digit (the "class" level) defines one of seven major enzymatic reaction types, e.g., hydrolases that use water to break a chemical bond. The second (the "subclass" level) and third digits (the "sub-subclass" level) represent more specific reaction types within their parent (sub-)class. The fourth digit (the "serial number" level) is a unique identifier assigned to a specific enzyme, usually determined by a specific substrate. For example, the enzyme number EC 1.3.8.2 represents an enzyme in class 1 (Oxidoreductases), subclass 3 (Acting on CH-CH group of donors), sub-subclass 8 (With a flavin as acceptor) with a unique number 2 (to identify a specific substrate 4,4'-diapophytoene).

\begin{table}[!ht]
    \centering
    \resizebox{0.65\columnwidth}{!}{
    \begin{tabular}{cc}
        \hline
        Level & \# classes\\
        \hline
        1 & 7\\
        2 & 24\\
        3 & 31\\
        4 & 462\\
        \hline
        \# classes in the dataset: & 4566 \\ 
        \hline
    \end{tabular}
    }
    \caption{\label{tab:classes_per_level}
    Number of classes per level in the ENZYME dataset
    }
\end{table}

\subsection{Construction}
The BRENDA database \cite{chang2021brenda} was parsed to collect enzymes with both EC number definitions and PubMed references. Following an inspection of retrieved entries, we downloaded the freely available PDFs for each of the articles from PubMed\footnote{These journal articles and preprints are accessible via Open Access that allows reuse.}. To extract well-formatted text from these files, we used an open-source tool called SciPDFParser\footnote{\url{https://github.com/titipata/scipdf_parser}}. For this work, we use the title, abstract and introduction sections from each paper. 

\section{Data Efficiency}

In Section \ref{data_efficiency}, we initially examined the performance of a restricted training set on the ENZYME dataset. In this subsequent section, we broaden our scope to encompass two additional datasets, namely WOS and NYT. Table \ref{tab:data_eff_ablation} shows the comparison between HBGL and HiGen for WOS and NYT for different proportions of training data. Within this experimental framework, our primary objective is to assess the performance of HiGen in comparison to the best-performing baseline model, HBGL. Our results show a noteworthy trend as the proportion of training data increases: there is a consistent and incremental improvement in performance for both HBGL and HiGen. However, what distinguishes our findings is the remarkable consistency with which HiGen outperforms HBGL across all fractions of training data, as reflected by superior Micro and Macro F1 scores.

\begin{table*}[ht]
\centering
\resizebox{\linewidth}{!}{%
\begin{tabular}{|c|cc|cc||cc|cc|}
\hline
 \multirow{3}{*}{Data Proportion}  & \multicolumn{4}{c||}{WOS} & \multicolumn{4}{c|}{NYT} \\
 \cline{2-9} 
 & \multicolumn{2}{c|}{HBGL}  & \multicolumn{2}{c||}{HiGen}  & \multicolumn{2}{c|}{HBGL}  & \multicolumn{2}{c|}{HiGen} \\
 \cline{2-9} 
 & Micro F1 & Macro F1 & Micro F1 & Macro F1 & Micro F1 & Macro F1 & Micro F1 & Macro F1 \\
\hline
0.1 & 82.35 & \textbf{75.01} & \textbf{82.92} & 73.83 & 74.90 & \textbf{57.96} & \textbf{74.95} & 54.96 \\
0.3 & 85.35 & 79.06 & \textbf{85.65} & \textbf{79.19} & \textbf{77.84} & 64.88 & 77.67 & \textbf{65.24} \\
0.5 & 86.11 & \textbf{80.30} & \textbf{86.63} & 79.96 & 78.75 & 66.75 & \textbf{79.57} & \textbf{69.36} \\
0.7 & 86.46 & 80.85 & \textbf{86.48} & \textbf{80.86} & 79.72 & 68.36 & \textbf{79.89} & \textbf{69.34} \\
\hline
\end{tabular}}
\caption{Performance on WOS and NYT datasets on varying the training data proportion for HBGL (left) and HiGen (right). We used HBGL because it is the best-performing model to compare against.}
\label{tab:data_eff_ablation}
\end{table*}

\section{Performance Analysis}
\label{sec:ablation_study_2}
\subsection{Ablation Study Results}
The performance figures for the ENZYME and NYT datasets shown in Tables \ref{tab:ablation_enz} \& \ref{tab:ablation_nyt} respectively, confirm the analysis presented in the previous section for the WOS dataset. The pretraining step is absolutely crucial as it gives the largest performance boost. Both the semantic ($\mathcal{L}_{S}$) and token ($\mathcal{L}_{T}$) losses significantly contribute to improving the model's Macro-F1 score while the contribution of the output space loss ($\mathcal{L}_{O}$) is significant but to a lesser extent. Across all scenarios, the removal of pretraining has the most profound impact, while omitting the proposed loss functions significantly influences performance, although with a less pronounced effect as compared to pretraining (Tables \ref{tab:ablation}, \ref{tab:ablation_enz} \& \ref{tab:ablation_nyt}).

Considering the improvement in HiGen over the Vanilla BART model, for the WOS dataset, the Mirco-F1 score improves by 1.13 points and the Macro-F1 score goes up by 2.11 points. For datasets with progressively larger hierarchies, NYT and ENZYME, we observe more pronounced improvements. In the case of NYT, which features a larger hierarchy than WOS with 166 unique labels and 8 levels, the Micro-F1 score improves by 0.81 points, and the Macro-F1 score shows a 3.11 point enhancement. The ENZYME dataset features a four-level deep hierarchy but with a notably higher count of unique labels (4566), surpassing both WOS and NYT. Here, the Micro-F1 score sees a substantial 4.5 points increase, while the Macro-F1 score remarkably jumps by 16.39 points.

Across all three datasets, a notable disparity is evident wherein Vanilla-BART exhibits subpar performance in comparison to the proposed HiGen model. This highlights the pivotal contribution of the proposed loss functions and the pretraining strategy in boosting the generative model’s performance. Specifically, in the Data Efficiency and Ablation Study analyses, we showcase the poor performance of Vanilla BART in contrast to HiGen.

\begin{table}[!ht]
    \centering
    \resizebox{0.8\linewidth}{!}{%
    \begin{tabular}{lcc}
    \hline
    Model & Micro-F1 & Macro-F1\\
    \hline
    Vanilla BART & 88.11 & 67.76 \\
    HiGen & \textbf{92.61} & \textbf{84.15} \\
    \hspace{2mm}w/o pretraining & 88.16 & 69.05 \\
    \hspace{2mm}w/o $\mathcal{L}_O$ & 92.33 & 83.00 \\
    \hspace{2mm}w/o $\mathcal{L}_T$ & 92.44 & 82.38 \\
    \hspace{2mm}w/o $\mathcal{L}_S$ & 92.36 & 82.43 \\
    \hline
\end{tabular}}
\caption{\label{tab:ablation_enz} Ablation study on HiGen for the ENZYME dataset.}
\end{table}

\begin{table}[!ht]
    \centering
    \resizebox{0.8\linewidth}{!}{%
    \begin{tabular}{lcc}
    \hline
    Model & Micro-F1 & Macro-F1\\
    \hline
    Vanilla BART & 80.08 & 69.30 \\
    HiGen & \textbf{80.89} & \textbf{72.41} \\
    \hspace{2mm}w/o pretraining & 77.89 & 69.65 \\
    \hspace{2mm}w/o $\mathcal{L}_O$ & 79.10 & 71.95 \\
    \hspace{2mm}w/o $\mathcal{L}_T$ & 79.31 & 71.85 \\
    \hspace{2mm}w/o $\mathcal{L}_S$ & - & - \\
    \hline
\end{tabular}}
\caption{\label{tab:ablation_nyt} Ablation study on HiGen for the NYT dataset.}
\end{table}

\subsection{Comparison with Baselines}

For all cases, the addition of the proposed framework over Vanilla BART yields substantial performance improvements. While the generative backbone contributes to performance enhancement to a certain degree when compared to baselines, a comprehensive evaluation against more recent baselines (HBGL, HGCLR, and T5-based baselines; refer to Table \ref{tab:comparison}) reveals that Vanilla BART alone falls short of surpassing them. Hence, the additional modifications proposed under HiGen are absolutely essential.

HiGen outshines baselines by large margins when the hierarchy size and data imbalance increases. This is evident for the NYT and ENZYME datasets. HiMatch \cite{chen-etal-2021-hierarchy} suffers a notable performance degradation, particularly in the Macro-F1 scores, where HiGen w/o pretraining outperforms it (Table \ref{tab:comparison} and Tables \ref{tab:ablation_enz} \& \ref{tab:ablation_nyt} for HiMatch and HiGen respectively). Note that these models employ complex architectures to encode the hierarchical structure whereas HiGen w/o pretraining uses an encoder-decoder framework without any context of the hierarchical structure. To compete with the more recent baselines, HGCLR \cite{wang-etal-2022-incorporating} and HBGL \cite{jiang2022exploiting}, we utilize pretraining to allow our model to grasp the hierarchical structure. Importantly, the pretraining approach we introduce is straightforward to implement and well-established. Simultaneously, it allows us to maintain a streamlined model architecture while achieving substantial performance gains. 

\subsection{Computational Cost}
\label{computational_cost}
To emphasize the simplicity and efficacy of our proposed architecture, we conduct a comparison between HiGen and the best-performing baseline, HBGL, in terms of the training times and computational resource utilization. Table \ref{tab:training_times} provides an overview of the training times for both models across all datasets. Both models were trained on a single NVIDIA RTX A5000 (24G) GPU.

The reported times reveal a stark efficiency in our proposed approach, where HiGen is approximately 10 times faster than HBGL, notwithstanding the notably simpler architecture employed. Most importantly, despite the simpler design and reduced training times, HiGen achieves superior performance as compared to the aforementioned baseline. 

\begin{table}[ht]
\centering
\begin{tabular}{|c|c|c|}
\hline
Dataset & HiGen (hrs) & HBGL (hrs) \\
\hline
ENZYME & 1.5 & 11.6 \\
WOS & 1.6 & 15.2 \\
NYT & 1.2 & 13.2 \\
\hline
\end{tabular}
\caption{Comparison of training times for HiGen and HBGL on all datasets.}
\label{tab:training_times}
\end{table}

\section{Hyperparameter Study}
\label{sec:hyperparam_study}
To study the influence of the loss-balancing factors $\lambda_1$, $\lambda_2$ \& $\lambda_3$, we conduct a hyperparameter study for the WOS and ENZYME datasets. The results are reported in Tables \ref{tab:hyperparam_study} \& \ref{tab:hyperparam_study_enzyme}. Our training and hyperparameter strategy was two-phased. Initially, we set the loss balancing factor for $L_S$ to 1 based on preliminary experiments. Subsequently, an extensive search was conducted for $\lambda_1$ and $\lambda_2$, with values ranging from 1 to 1e-8 with a step factor of 0.1. From the results (Table \ref{tab:hyperparam_study}), we observed that our model was quite robust to changes in middle ranges. Notably, a distinct performance peak emerged at $\lambda_1 = 1e-3$ and $\lambda_2 = 1e-5$. 

\begin{table}
    \centering
    \resizebox{0.8\linewidth}{!}{
    \begin{tabular}{ccccc}
    \hline
    $\lambda_1$ & $\lambda_2$ & $\lambda_3$ & Micro-F1 & Macro-F1\\
    \hline
    1e - 2 & 1e - 2 & 1 & 86.63 & 79.18 \\
    1e - 3 & 1e - 3 & 1 & 87.13 & 81.03 \\
    \textbf{1e - 3} & \textbf{1e - 5} & \textbf{1} & \textbf{87.39} & \textbf{81.45} \\
    1e - 4 & 1e - 4 & 1 & 86.91 & 80.67 \\
    1e - 5 & 1e - 5 & 1 & 86.71 & 80.27 \\
    \hline
\end{tabular}}
\caption{\label{tab:hyperparam_study}Hyperparameter study on the WOS dataset.}
\end{table}

\begin{table}[t]
    \centering
    \resizebox{0.8\linewidth}{!}{
    \begin{tabular}{ccccc}
    \hline
    $\lambda_1$ & $\lambda_2$ & $\lambda_3$ & Micro-F1 & Macro-F1\\
    \hline
    1e - 2 & 1e - 2 & 1 & 92.31 & 80.06 \\
    1e - 3 & 1e - 3 & 1 & 92.39 & 82.81 \\
    1e - 3 & 1e - 5 & 1 & 92.38 & 83.81 \\
    \textbf{1e - 3} & \textbf{1e - 6} & \textbf{1} & \textbf{92.61} & \textbf{84.15} \\
    1e - 4 & 1e - 6 & 1 & 92.58 & 83.87 \\
    1e - 6 & 1e - 6 & 1 & 92.34 & 82.58 \\
    \hline
\end{tabular}}
\caption{\label{tab:hyperparam_study_enzyme}Hyperparameter study on the ENZYME dataset.}
\end{table}

To verify the generalizability of these findings, we performed a similar analysis for the ENZYME dataset. The performance figures have been reported in Table \ref{tab:hyperparam_study_enzyme}. The performance peaks around the same hyperparameter settings as observed for the WOS dataset. This pattern similarly holds true for the NYT dataset, where the optimal hyperparameter settings align with those of WOS. This coherence in hyperparameter settings greatly simplified the process of tuning multiple parameters. Since almost the same settings work across all datasets, we argue that our model is quite robust to these values and future works could use this setting as a good starting point on other datasets as well.

\section{Generating Pretraining Data}
\label{sec:chatgpt_pretraining}

Given the unavailability of similar datasets following the same label hierarchy for the WOS dataset, we employ the powerful ChatGPT API to generate the pretraining data. To capture the domain of science and the corresponding sub-category, we carefully designed the prompts. For instance, when focusing on the domain of "\textit{Mechanical engineering}" and sub-category "\textit{computer-aided design}", our curated prompt is as follows:

\begin{quote}
Write 20 different abstracts for scientific articles in the \textit{Mechanical Engineering} domain and \textit{computer-aided design}.    
\end{quote}

We modify the prompt accordingly for different domains and sub-categories. To ensure diversity and non-redundancy, we execute the API multiple times for each domain and sub-category combination, resulting in a well-balanced corpus of nearly 3000 abstracts. Refer to Table \ref{tab:chatgpt_eg} for some examples of abstracts generated by ChatGPT.

To give an idea about the dissimilarity between the original test set in the WOS dataset and the generated pretraining data, we find the overlap between them. In order to do so, we preprocess the abstracts from both data sources by removing the stop words. Then, we compare the Jaccard Similarity between the two sets of data. Averaging the Jaccard Similarity across all categories we get the value to be 4.12\%. This indicates that there is a minuscule overlap between the words (none of the overlaps exceeds 6\%) from the test set of WOS and the pretraining data obtained from ChatGPT. This indicates that there is no data leakage during the pretraining step.

\begin{table*}
    \centering
    \resizebox{\linewidth}{!}{%
    \begin{tabular}{|p{0.3\textwidth}|p{0.3\textwidth}|p{0.3\textwidth}|}
    \hline
    \hfil \textbf{Category} & \hfil \textbf{Sub-domain} & \hfil \textbf{Text}  \\ 
    \hline
    \hfil Computer Science & \hfil Data Structures & \small In this study, we propose a new data structure called the Bloom filter tree. The Bloom filter tree combines the space efficiency of Bloom filters with the fast search capabilities of trees. We show that the Bloom filter tree outperforms other data structures in terms of space usage and query time. \\ 
    \hline
    \hfil Civil Engineering & \hfil Stealth Technology & \small This article presents an experimental study on the effect of moisture on the mechanical properties of asphalt concrete. The study was conducted through laboratory tests on asphalt concrete samples with varying moisture content. The results show that moisture has a significant impact on the stiffness and strength of the asphalt concrete. \\ 
    \hline
    \hfil Biochemistry & \hfil Genetics & \small This article discusses the emerging field of microRNA research and its implications for gene regulation and disease pathogenesis. The research team investigates the role of microRNAs in various biological processes, including development, cell differentiation, and immune response. The findings highlight the potential of microRNAs as therapeutic targets for various diseases. \\ 
    \hline
    \end{tabular}}
    \caption{\label{tab:chatgpt_eg}Examples of abstracts generated by ChatGPT.
    }
    \vspace{-0.5em}
    
\end{table*}

\end{document}